\documentclass[10pt,twocolumn,letterpaper]{article}

\usepackage{wacv}              % To produce the CAMERA-READY version
\usepackage{graphicx}
\usepackage{amsmath}
\usepackage{amssymb}
\usepackage{booktabs}
\usepackage{xcolor}

\definecolor{darkgreen}{rgb}{0.0, 0.75, 0.0}
\newcommand{\better}[1]{\textcolor{darkgreen}{#1}}

\usepackage[pagebackref,breaklinks,colorlinks]{hyperref}

\usepackage[capitalize]{cleveref}
\crefname{section}{Sec.}{Secs.}
\Crefname{section}{Section}{Sections}
\Crefname{table}{Table}{Tables}
\crefname{table}{Tab.}{Tabs.}

\def\wacvPaperID{*****} % *** Enter the WACV Paper ID here
\def\confName{WACV}
\def\confYear{2025}

\begin{document}

\title{
% Mask Diffuser: Boosting Image-to-Image Synthesis with Prism 
% Boosting Image-to-Image Consistency with High-Entropy Diffusion
Diffusion Prism: Enhancing Diversity and Morphology Consistency in Mask-to-Image Diffusion
}

\author{Hao Wang$^1$, Xiwen Chen$^1$, Ashish Bastola$^1$, Jiayou Qin$^2$, and Abolfazl Razi$^1$\thanks{Corresponding author}\\
$^1$Clemson University\\
$^2$Stevens Institute of Technology\\
{\tt\small \{hao9, xiwenc, abastol\}@g.clemson.edu, jqin6@stevens.edu, arazi@clemson.edu}
% \and
% Name \\
% Stevens Institute of Technology\\
}
\maketitle

% \xc{After I read the paper, I think it is better to make the paper specific to dendrites. You can say like the dendrite is low-entropy, current diffusion models are challenging to handle them. Then, we solve it, xxxxx. Otherwise, this claim is too general, and the experiment is not convincing. I also think this paper needs more revision before submission. Otherwise, it has a low chance to get accepted. }

%%%%%%%%% ABSTRACT
\begin{abstract}
The emergence of generative AI and controllable diffusion has made image-to-image synthesis increasingly practical and efficient. However, when input images exhibit low entropy and sparsity, the inherent characteristics of diffusion models often result in limited diversity. This constraint significantly interferes with data augmentation in many fields.
To address this, we propose \textbf{Diffusion Prism}, a training-free framework that efficiently transforms binary masks into realistic and diverse samples while preserving morphological features. We explored that a small amount of artificial noise will significantly assist the image-denoising process. To prove this novel mask-to-image concept, we use nano-dendritic patterns as an example to demonstrate the merit of our method compared to existing controllable diffusion models. Furthermore, we extend the proposed framework to other biological patterns, highlighting its potential applications across various fields. 
Our source code and sample datasets are available at: \url{https://arazi2.github.io/aisends.github.io/project/Prism}
% \url{}

\vspace{-0.5cm}
\end{abstract}

% 1. concept draw 
% 2. experiments
% 3. equations 
% 4. text/citations
% 5. figures

%%%%%%%%% BODY TEXT
\section{Introduction}
\label{sec:intro}

Image quality plays an important role in various data-scarce domains. For instance, research in biometrics, material science, and medical imaging \cite{minaee2023biometrics,ma2024segment} heavily relies on high-quality raw data, precise annotation, and effective data augmentation to overcome the lack of sufficient datasets \cite{1282003,lahiri2020retinal,ozbey2023unsupervised}. 
With deep learning dominant in recent research, high-quality data pre-processing and efficient utilization lay a solid foundation for downstream tasks such as segmentation, detection, recognition, and classification \cite{wu2019u,bonaldi2023deep,kirillov2023segment,zhu2023beyond,wu2024medsegdiff}.

\begin{figure}[htbp]
    \centering
    \centerline{\includegraphics[width=1\columnwidth]{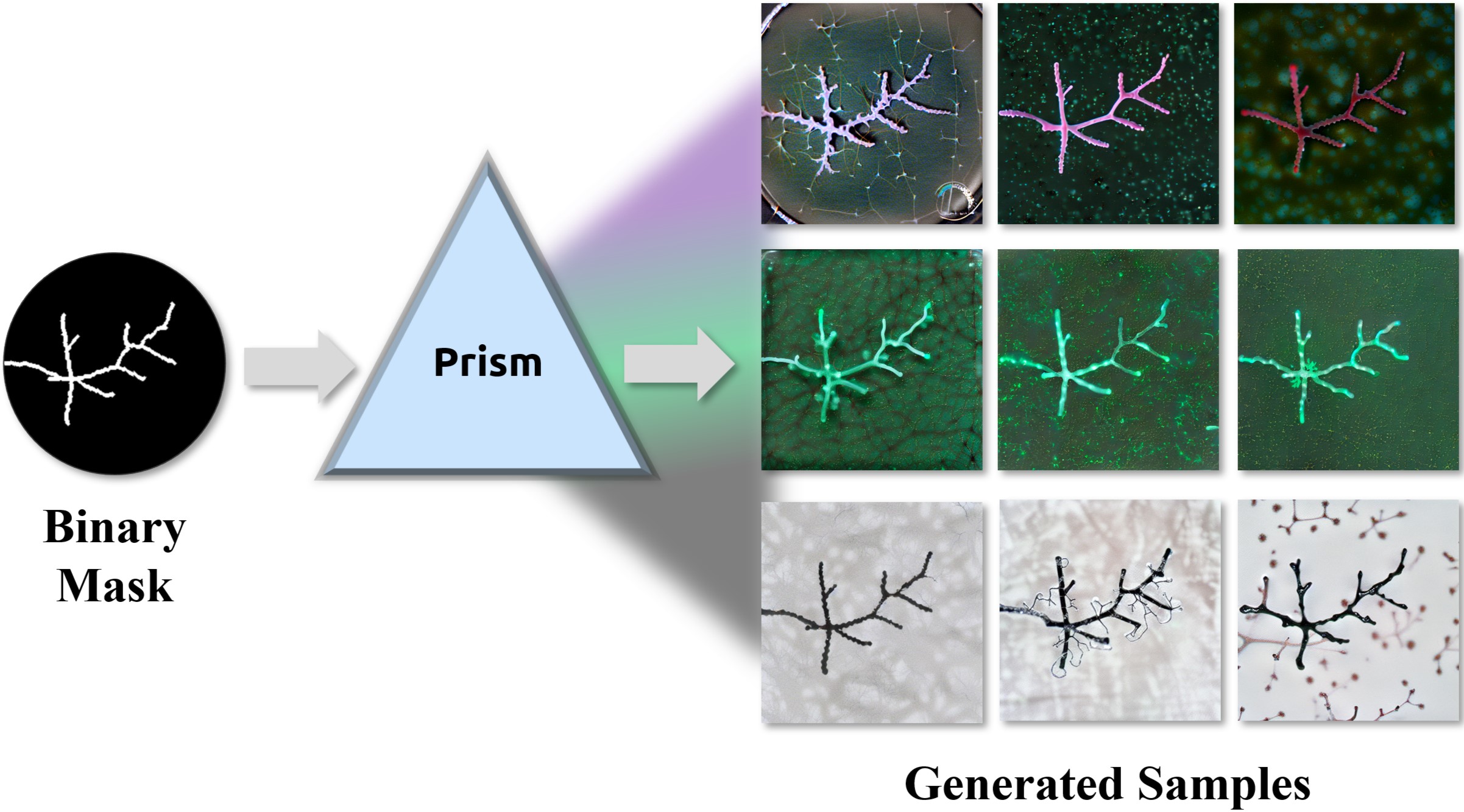}}
    \caption{Concept of the Diffusion Prism}
    \label{fig:teaser}
    \vspace{-0.5cm}
\end{figure}

\begin{figure*}[htbp]
    \centering
    \centerline{\includegraphics[width=1\textwidth]{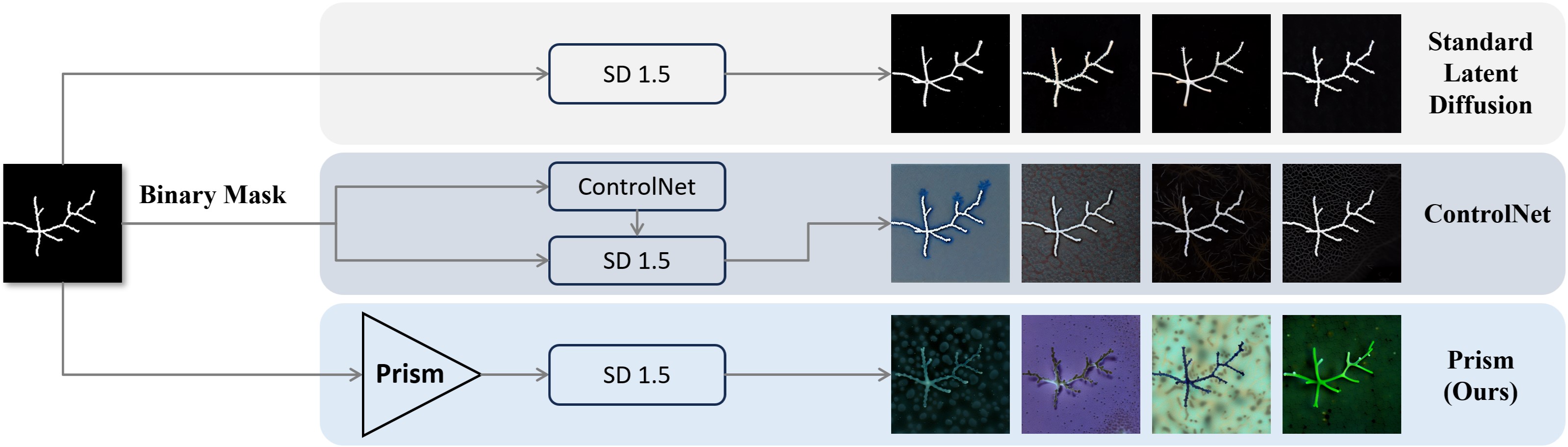}}
    \caption{Comparison of different diffusion frameworks. The proposed method Prism is an individual add-on module that does not require any training and does not interact with the vanilla diffusion model}
    \label{fig:control}
    \vspace{-0.25cm}
\end{figure*}

Recently, generative models such as GANs (Generative Adversarial Networks), have demonstrated the ability to generate realistic images for data synthesis \cite{wu2019u,zhu2023otre,melnik2024face}, enabling researchers to expand datasets with minimal manual effort. More approaches such as style transfer, have also been widely used to transform existing images into new ones with specified styles \cite{holden2017fast,wright2022artfid}. However, these methods exhibit inherent limitations: their reliance on predefined style templates constrains their ability to generate truly novel and diverse samples, and their need for computationally expensive fine-tuning or retraining reduces their practicality \cite{torkzadehmahani2019dp,loey2020deep}. 
% For complex and high-dimensional patterns like dendritic structures, these shortcomings make traditional approaches insufficient for achieving the level of diversity and richness required in scientific and security applications.
 
Latent diffusion models have rapidly become the dominant framework for high-quality image generation across various computer vision tasks \cite{rombach2022high}. Using textual prompts allows the synthesis of semantically rich and visually detailed images \cite{choi2023custom,mokady2023null}, with applications spanning digital arts and content creation \cite{meng2021sdedit, zhou2023denoising}. Recent research further introduced controllable generation frameworks, such as ControlNet and Uni-ControlNet \cite{zhang2023adding,zhao2024uni}, that enhance the flexibility of diffusion models by incorporating input-specific images (pose map, depth map, etc.) as structural guides. These developments have significantly improved the precision and applicability of AI-generated content, especially in fields such as design and manufacturing \cite{zhang2024transparent}. 

% Despite the progress in conditional diffusion, these methods are far from flawless. The final quality of the generated image often mirrors the limitations of the initial input. Sparse inputs, such as binary masks with minimal detail, lead to outputs with simplistic structures, bland backgrounds, and lacking of complexity. This reliance on the initial input highlights the need for methods to address such constraints.

While these frameworks excel in entertainment and creative industries, the gap still exists between their practical application and conventional scientific research such as material science and medical imaging \cite{lin2021temimagenet,lee2024microstructure,wang2024rbad}. In these fields, the ability to generate realistic and diverse samples from sparse binary masks could enable transformative advances in data augmentation and morphological analysis \cite{ozbey2023unsupervised}. However, we identify a major limitation in existing controllable diffusion methods: lack of image diversity when working with sparse inputs. These challenges arise from the denoising process, where sparse inputs fail to guide the generation of rich and complex outputs.

% This paper introduces Diffusion Prism, a method designed to enhance both the richness and structural consistency of outputs without requiring additional model training. By embedding subtle artificial signals into sparse input images, Diffusion Prism enables the generation of complex and detailed outputs. These signals influence the denoising process, breaking through the barriers imposed by sparse inputs. Using dendritic patterns as a study case, this paper investigates the role of initial signals in shaping output complexity and alignment with textual guidance.

% Experimental evaluations demonstrate the improvements in dataset diversity facilitated by Diffusion Prism. Additionally, the analysis explores how noise and chromatic alterations influence denoising behavior. A segmentation approach based on random forests is applied to verify the method’s ability to maintain structural integrity. Finally, the paper extends its findings to other domains, illustrating the broader applicability of this approach. 

% Prism
To address the challenge of generating realistic samples from binary masks, we propose the \textbf{Diffusion Prism}, a simple yet efficient method that requires no additional training or fine-tuning of the diffusion model. 
The Diffusion Prism is an individual add-on module that is based on the pre-trained Stable Diffusion v1.5 (SD1.5) \cite{rombach2022high} and only modulates the input images in the pixel space to achieve domain shift without altering the diffusion model’s parameters, similar to light refraction through a prism in the real world, as shown in Figure \ref{fig:teaser}.
The Diffusion Prism enhances both the diversity and texture of generated images with morphological consistency by introducing controllable noise and chromatic aberration into the input. 
For convenience, we use the term \textbf{Prism} to represent this proposed method in the rest of the paper.  
% The result is a versatile image generation technique that can be applied to various fields, where both morphological structure and diversity are critical.

Our contribution can be summarized as:
\begin{itemize}
    \item Training-Free Framework: Our proposed method requires no additional training or fine-tuning.
    \item Morphology Consistency: We employ a lower denoising strength to preserve the morphological information from the input manually.
    \item     Enhanced Diversity: By manipulating the input image in the pixel domain, it effectively expands the diversity of input images.
\end{itemize}

To prove this novel mask-to-image concept, we employ the nano-dendritic patterns $-$ a mathematically generated high-entropy random patterns $-$ as an example in our experiments to showcase the feature of our method.
The results demonstrate that the proposed Prism can significantly improve the image diversity of dendritic samples, while not sacrificing the integrity of morphology structures of the input binary mask.

% The versatility and effectiveness of our approach offer a practical solution for data augmentation and other applications.

% \begin{figure}[htbp]
%     \centering
%     \centerline{\includegraphics[width=1\columnwidth]{Figure/cover.png}}
%     \caption{Sample images from different diffusion frameworks.}
%     \label{fig:diffuser}
% \end{figure}

\section{Related Work}

\subsection{Dendritic Patterns}
\begin{figure}[htbp]
    \centering
    \centerline{\includegraphics[width=1\columnwidth]{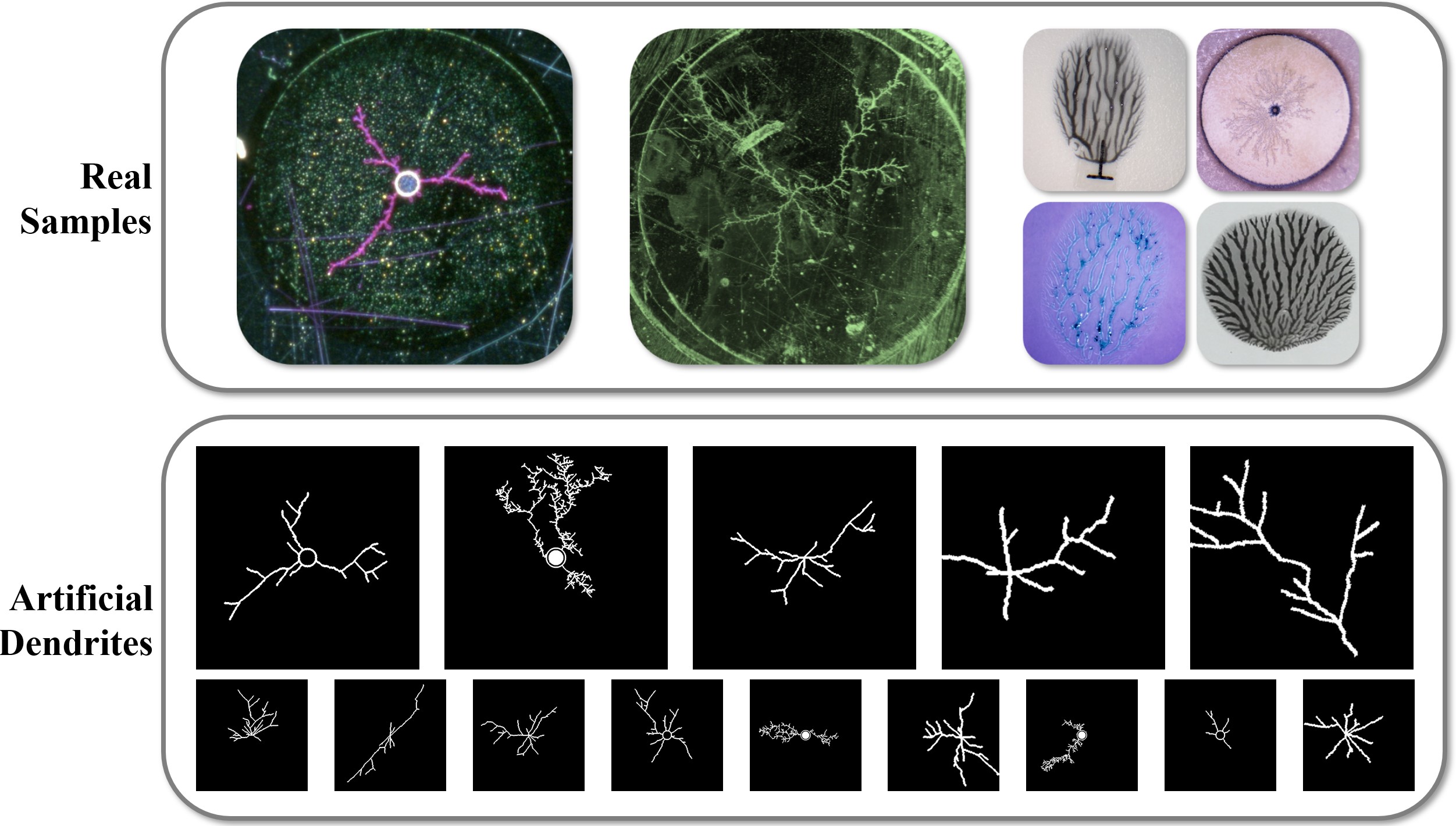}}
    \caption{Dendrite samples. Real samples (upper) are taken in lab microscopes; Artificial samples (lower) are generated using mathematical algorithms}
    \label{fig:dendrite}
\end{figure}

Dendritic patterns, a class of nano-scale digital tokens, have emerged as highly representative structures for studying and advancing biometric and security applications \cite{10216773,fang2024fairness}. These patterns, formed through stochastic natural processes, are characterized by their distinctive morphological features, high entropy, and randomness \cite{cuntz2007optimization,spruston2008pyramidal}. Unlike traditional biometric traits, dendritic patterns can encode vast amounts of multidimensional information while remaining resistant to reverse engineering and prediction \cite{chen2023dh,chen2024enhancing}. Furthermore, their structural complexity can be mathematically modeled, providing researchers with an ideal tool for exploring the intricate characteristics of biological information and developing novel security frameworks \cite{10216721}, as shown in Figure \ref{fig:dendrite}.
% However, the scarcity of such naturally occurring patterns poses a significant limitation, restricting their use in large-scale applications and preventing comprehensive studies of their potential. 

However, the lack of annotated datasets for these structures presents a major challenge, as capturing their intricate geometry and fine details is both labor-intensive and costly \cite{lin2021temimagenet}. 
Thus, a method to generate mask-to-image pairs is valuable for downstream tasks including segmentation, object recognition, and skeleton extraction, as the masks serve as ground truth for the generated images \cite{kirillov2023segment, zhou2023maskdiffusion, kawano2024maskdiffusion,wang2024flame}. 

% To address this, we leverage pre-trained diffusion models to generate synthetic images of tree-like structures, allowing us to bypass the scarcity of real-world samples and represent their complex morphology.
% Moreover, this method has broad applications in other domains, such as industrial design and augmented reality \cite{}, where generating structured images can reduce the time and cost of manual data annotation. By synthesizing images with predefined structures, we provide an efficient solution for large-scale data generation and annotation.

\subsection{Image-to-Image Diffusion}

A standard latent diffusion model consists of three key components: the Variational Autoencoder (VAE), which compresses high-dimensional input images into a lower-dimensional latent space to reduce computational complexity; the denoising U-Net, which refines noisy latent representations into coherent outputs; and a CLIP-based tokenizer that bridges the gap between textual prompts and images \cite{radford2021learning}. Together, these components enable the efficient synthesis of semantically rich and visually detailed images \cite{rombach2022high}.

As the demand for more control over the generation process grew, models such as ControlNet introduced the controllable image-to-image diffusion, allowing precise control over generated content by incorporating structural guides such as masks and sketches \cite{zhang2023adding, zhao2024uni}. These methods have proven effective in industrial applications, where generated content must meet specific structural and stylistic requirements \cite{zhang2024transparent}.

Despite their success, existing controllable diffusion frameworks face critical limitations when applied to sparse inputs. As shown in Figure \ref{fig:control}, sparse binary masks fail to guide the generation of wealthy and diverse content. This is partially because, in controllable diffusion architecture, the controlling network mostly guides the high-frequency signals from the conditional images, thus preserving the morphology from the input \cite{zhao2024uni}. However, due to a strong relation between the initial image and the final decoded tensor, a sparse input can hardly generate high-entropy information by only relying on the denoising sampler \cite{everaert2024exploiting}.

% Furthermore, under strong noise conditions, high-frequency details are lost during the denoising process, leading to outputs that lack the richness and complexity required for scientific and industrial applications.

% \begin{figure*}[htbp]
%     \centering
%     \centerline{\includegraphics[width=0.7\linewidth]{Figure/frame.png}}
%     \caption{The concept of Mask Diffuser: synthesize diverse samples using binary masks.}
%     \label{fig:concept}
% \end{figure*}

\begin{figure*}[htbp]
    \centering
    \centerline{\includegraphics[width=0.8\textwidth]{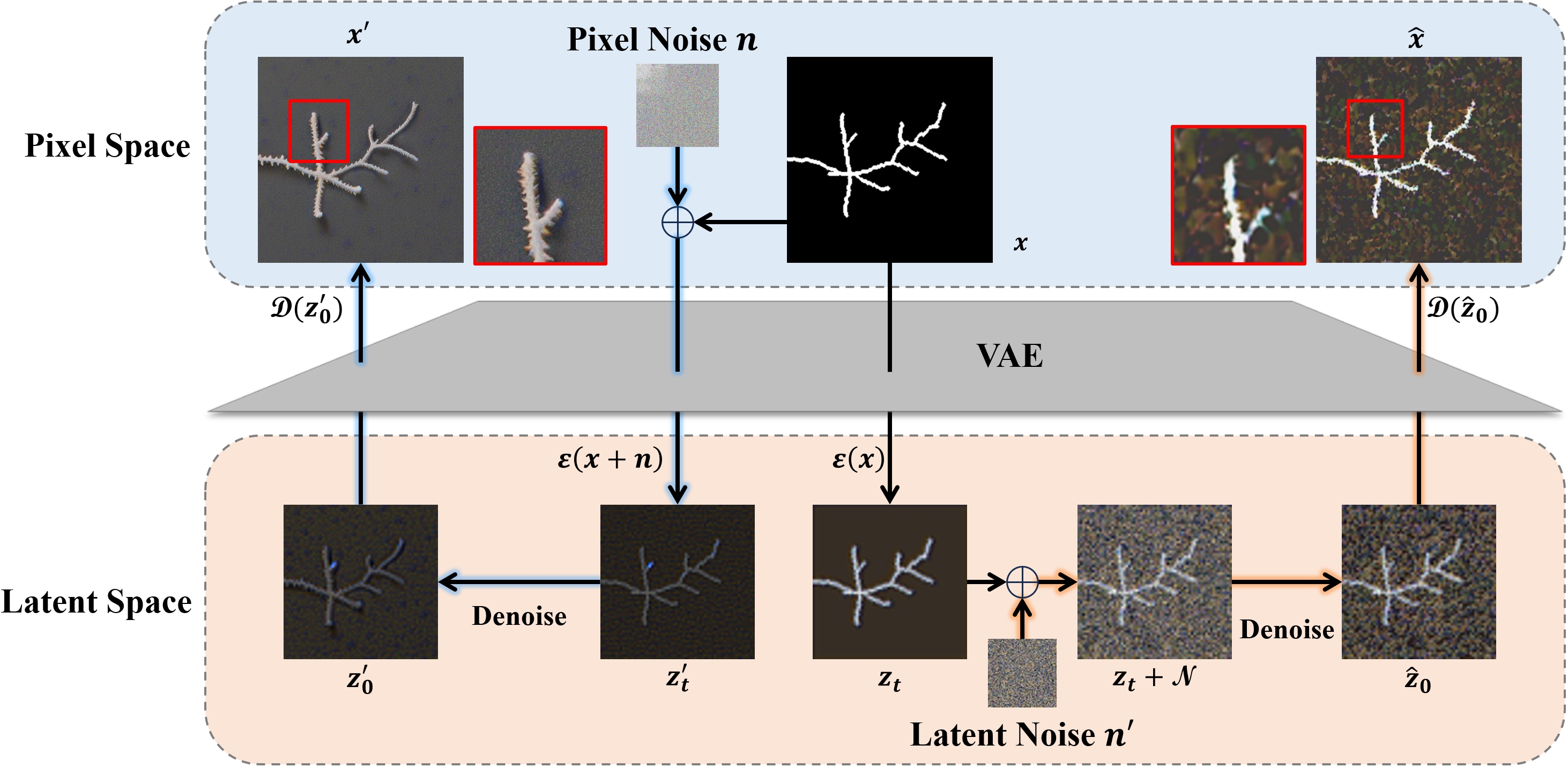}}
    \caption{Noise injection comparison: pixel space vs. latent space.}
    \label{fig:latent}
\vspace{-0.25cm}
\end{figure*}

\section{Methodology}
\label{sec:eq}
% Enhancing the diversity of dendritic patterns while preserving the morphological integrity of the input images is a challenge for existing image-to-image translation models. 
The sparse and low-entropy nature of binary masks makes the denoising process difficult, as the randomness introduced during diffusion is often insufficient to generate diverse backgrounds. 
To address this, we proposed our method based on a pre-trained Stable Diffusion v1.5 to explore the image denoising process.

\subsection{Signal Degradation in Denoising}

\begin{figure}[htbp]
    \centering
    \centerline{\includegraphics[width=1\columnwidth]{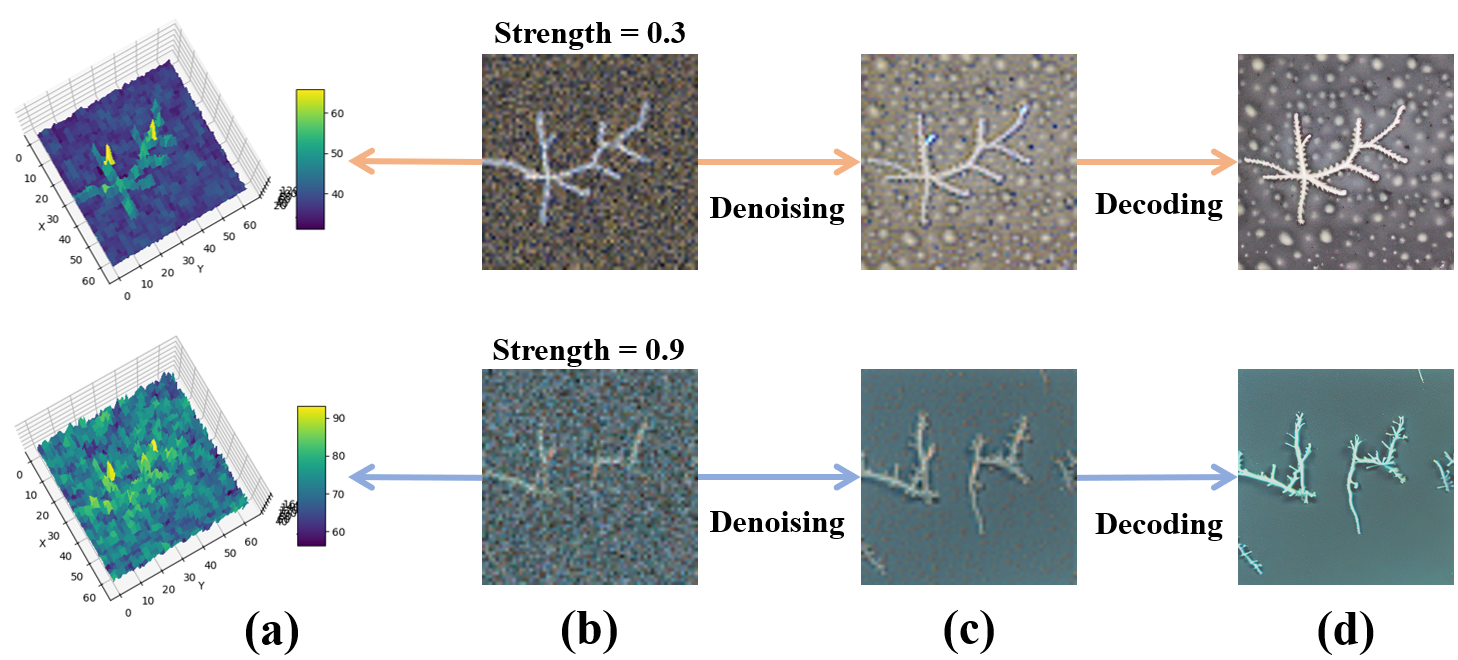}}
    \caption{Noise sampling schedule in image-to-image synthesis.}
    \label{fig:sampling}
    \vspace{-0.25cm}
\end{figure}

The uncertainty introduced by high-denoising strength is a key challenge in image-to-image translation. Although high denoising strength increases diversity, it degrades high-frequency information, removing important information that may contain morphological details. This causes parts of the input image in the latent space to be treated as noise, making it difficult to maintain morphological consistency with the input, as illustrated in Figure \ref{fig:sampling}. 

To examine the details, we formulate the problem by using the vanilla DDIM sampler from latent diffusion \cite{song2020denoising,rombach2022high}.  The initial latent tensor $z$ is generated during forward diffusion using the following equation:
\begin{align}
\label{eq:e}
    z_t = \alpha_t \cdot x_0 + \sqrt{1 - \alpha_t} \cdot \epsilon.
\end{align}

% \xc{Double check this claim. This update rule is from the training phase. With any scheduler of $\alpha$, it will finally become a random image. }
Here, \( \alpha_t \) controls how much of the original image \( x_0 \) is retained over time \( t \), while \( \epsilon \), sampled from a Gaussian distribution \( \epsilon \sim \mathcal{N}(0, 1) \), represents the latent noise added to obscure the image. 
In the denoising (reverse diffusion) process, as \( t \) increases, the denoise sampler will predict a higher amount of noise at the initial step, as shown in Figure \ref{fig:sampling}(a)(b). 
As a consequence, the influence of the input image diminishes, and noise becomes more dominant, leading to signal degradation and loss of high-frequency details, as shown in Figure \ref{fig:sampling}. 
This explains the phenomenon that increasing the denoising strength introduces more randomness but also increases the risk of losing critical information from the input image. 
To balance morphology consistency and diversity, a lower denoising strength is often recommended. 

% \hw{
% After extensive testing, we set the denoising \textbf{strength} to $0.3$, which we found to be the optimal balance for preserving the original morphology while still introducing sufficient diversity. This value is consistently used across all experiments to ensure fair comparisons with other methods.
% }
% generated images retain structural integrity while still achieving sufficient diversity.

\subsection{Initial Entropy}
When using lower strength during denoising, we found that binary input images, such as masks, fail to produce diverse outputs. Such low-entropy images often produce minimal visual content because they lack sufficient randomness. This also explains why mainstream text-to-image models start with random matrices to provide the necessary diversity for effective image generation \cite{rombach2022high}.

We found that introducing artificial signals into the initial image can significantly influence the denoising process. These artificial signals will be misinterpreted as noise, which affects the prediction of noise and the calculation of the noiseless tensor \(z_0\). 
Specifically, the process can be expressed as: 
\begin{align}
    z_t' = \mathcal{E}(x+{n}) \quad 
\end{align}
\begin{align}
    \epsilon_\theta(z_t',t) = \epsilon_\theta(z_t,t) + \delta 
\end{align}
Here, $\mathcal{E}$ is the encoder of the pre-trained VAE, ${n}$ represents the introduced signal, $\epsilon_\theta$ is the predicted noise at step $t$, and $\delta$ is the residual term at step $t$ that is introduced by the signal $n$ during encoding. As the predicted noise changes, the noiseless tensor $z_0$ also alters:
\begin{align}
z'_0 = \frac{z_t - \sqrt{1 - \alpha_t} (\epsilon_\theta(z_t,t) + \delta)}{\sqrt{\alpha_t}} \quad ,
\end{align}
\begin{align}
z'_0 = & z_0 - \frac{\sqrt{1 - \alpha_t}}{\sqrt{\alpha_t}} \delta \\ \nonumber
                     = & z_0 - \hat{\alpha} \delta. \quad
\end{align}
Here, the term $\hat{\alpha} \delta$ represents the introduced information in the latent space. 
By adding \(n \) to the initial image, the additional signal becomes part of the image content rather than being treated as noise due to the domain shift, as shown in Figure \ref{fig:latent}. This ensures the consistency of the denoising steps while enriching the content, resulting in more detailed final images.

\begin{figure*}[htbp]
    \centering
    \centerline{\includegraphics[width=1\linewidth]{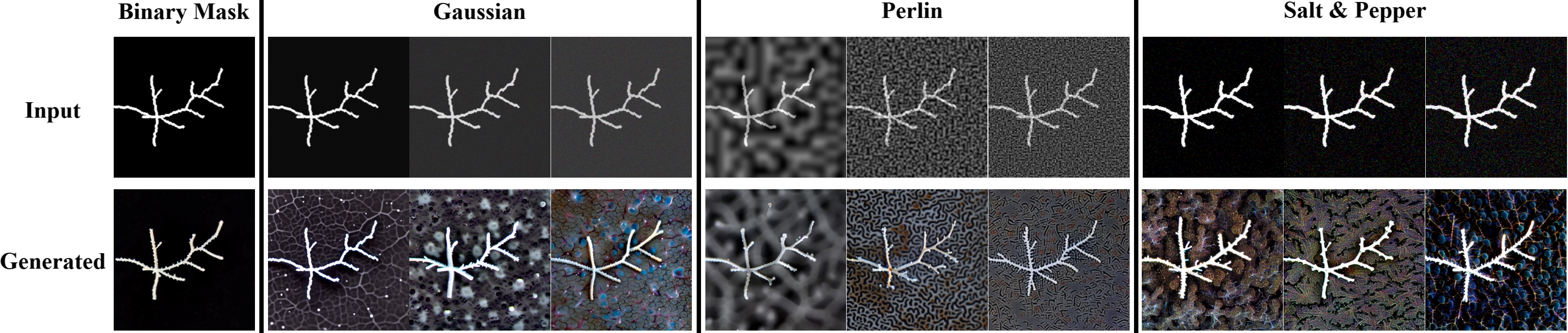}}
    \caption{Noise type comparison, the input are binary mask with different types of noise, and the outputs are the generated images with the proposed method.}
    \label{fig:type}
    \vspace{-0.25cm}
\end{figure*}

\subsection{Domain Shift in Diffusion}
\label{sec:domain}
% From a domain transfer perspective, the output image domain is primarily determined by the \textbf{initial tensor}—a combination of the input image and the sampled noise. 
There exist two possible ways to achieve domain shift in diffusion: pixel-level operations and latent space operations \cite{everaert2024exploiting}. 
However, due to the different characteristics of the image in the pixel domain versus the latent space, manipulating in latent space is risky. 
For instance, the latent diffusion process can be simplified as:
\begin{align}
    \hat{x} = \mathcal{D}(z) = \mathcal{D}(\mathcal{E}(x)), \quad 
\end{align}
where $x$ is the input image, $\mathcal{E}$ and $\mathcal{D}$ are the encoder and decoder of the pre-trained VAE, $z$ is the latent tensor, and $\hat{x}$ is the denoised image \cite{rombach2022high}.

Thus, making any changes directly in the latent space tends to result in corrupted images in the pixel domain. As shown in Figure \ref{fig:latent}, altering the content of the latent tensor may shatter the representation of the image in latent space.  Additionally, to align the latent tensor with the intended transformation, retraining or fine-tuning the model is often required \cite{everaert2023diffusion}.
In contrast, the translation from pixel to latent space effectively projects the pixel distribution into the latent distribution. This process ensures that all information present in the pixel space is coupled into the latent space, making it easier to fuse with text features.

% In this work, we opt for \textbf{pixel-level manipulation} for minimal and more effective transformation. 

\subsection{Diffusion Prism}
As proved in Section \ref{sec:domain}, latent operation poses a high risk of domain collapse. Thus, in this paper, we employ image manipulation in the pixel domain, which allows for minimal yet effective transformations by introducing artificial signals or noise.
In addition to blending noise into the initial image, we also introduce \textit{chromatic aberration} as an additional form of domain warping. This process involves shuffling pixel values across channels, which simulates optical distortion without compromising the structural integrity of the input image. 
% By doing so, we induced a controlled domain shift and preserved the overall morphology.

When using Prism in practical applications, we typically derive the mean and standard deviation from a reference image to emulate a specific style, as shown in Figure \ref{fig:teaser}. However, in the experiments, we use a random matrix as the sample image $I$ to ensure maximum diversity. 
Specifically, we evaluate the pixel distribution in each color channel by calculating:
\begin{align}
\mu = \frac{1}{N} \sum_{x,y} I(x,y),
\end{align}
\begin{align}
\sigma = \sqrt{\frac{1}{N} \sum_{x,y} (I(x,y) - \mu)^2},
\end{align}
where \( \mu \) and \( \sigma \) are the mean and standard deviation of the pixel values, \( N \) is the total number of pixels in a sample image \(I\), and \( I(x,y) \) is the pixel value at position \( (x,y) \). 
Furthermore, we incorporate the original color information with \( \mu \) and \( \sigma \) into the mask $M$, along with artificial noise $n$:
\begin{align}
        n \sim \mathcal{N}(\mu, \sigma)
\end{align}
\begin{align}
        M(x, y) =& (M(x, y)\cdot \sigma + \mu )+ n
\end{align}
This process introduces controlled variations via samples from a standard normal distribution. Note that this normal distribution is different than the denoise sampler, as mentioned in Equation \ref{eq:e}. 
This approach ensures that the generated image has both structural integrity from the mask and sufficient visual variability from the added artificial noise.

% This combination enriches the input entropy while ensuring color consistency, enabling the generation of diverse yet morphologically consistent outputs. 
% This process ensures that the resulting image retains structural consistency from the mask while introducing enough randomness and noise to generate visually rich and diverse textures, improving the overall visual realism.

% This method ensures that the generated images maintain both their morphological integrity and exhibit a rich diversity in texture while avoiding the complexities associated with latent space manipulation and model retraining.

\begin{figure}[htbp]
    \centering
    \centerline{\includegraphics[width=1\linewidth]{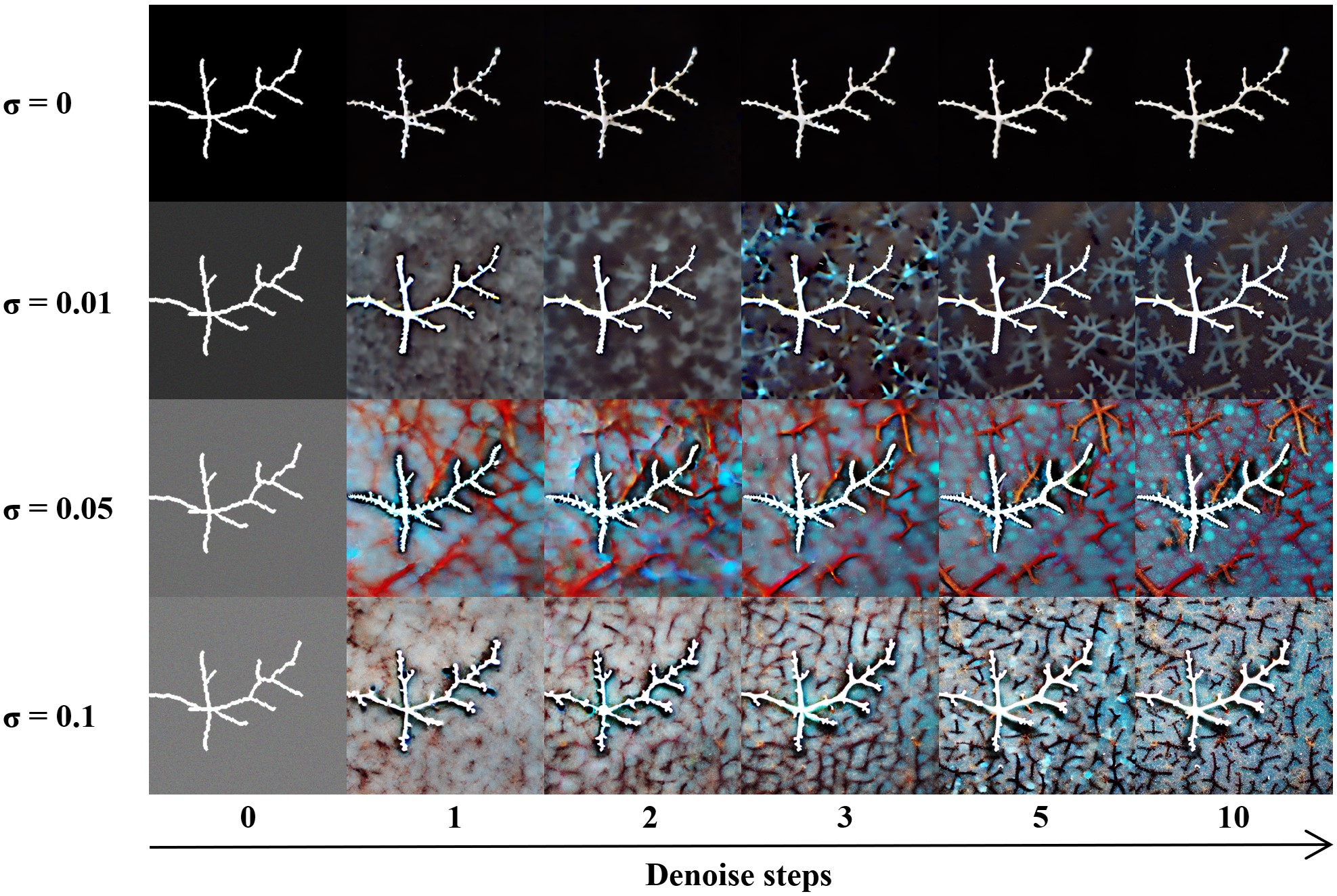}}
    \caption{Mask diffusion process: adding Gaussian noise into the input image can increase the richness of the background. Denoising step: $10$, text prompt scale: $10$, denoising strength: $0.3$, text prompt: \textit{"a dendrite sample"}.}
    \label{fig:noise}
    \vspace{-0.25cm}
\end{figure}

\begin{figure*}[htbp]
    \centering
    \centerline{\includegraphics[width=1\linewidth]{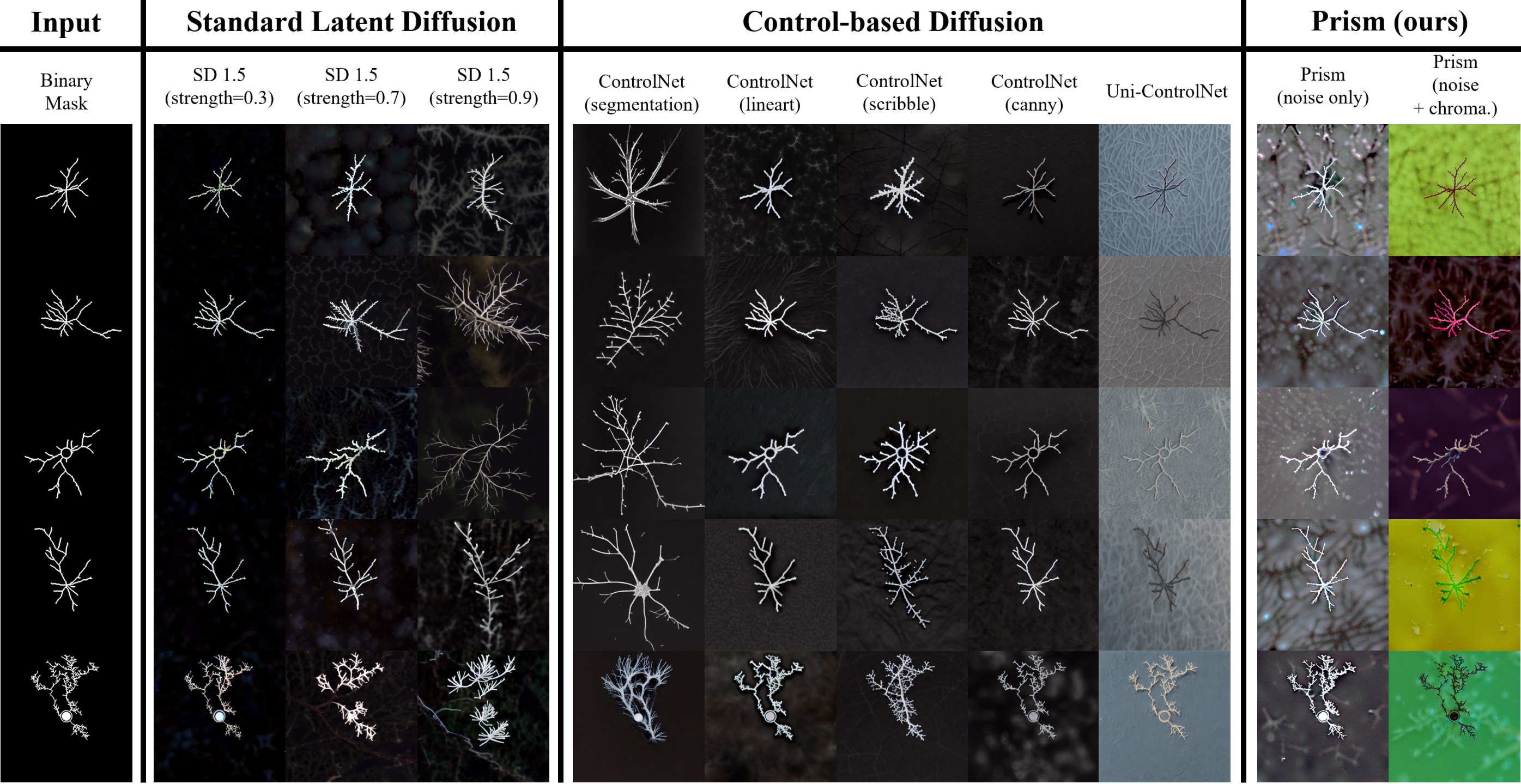}}
    \caption{Qualitative test. We use three different diffusion frameworks (standard latent diffusion, control-based diffusion, and our method) to obtain sample datasets. The input is a binary mask and all frameworks use image-to-image diffusion to generate samples.}
    \label{fig:grid}
    \vspace{-0.25cm}
\end{figure*}

\subsection{Impact of Noise in Prism}

% We evaluate how adding artificial noise to the initial image affects the generated image's diversity and background details. Gaussian noise with varying standard deviation \( \sigma \) was introduced at the pixel level. As shown in Figure \ref{fig:noise}, , and increasing the noise further enhances the variation in the background.
% Figure \ref{fig:noise} shows that adding noise significantly impacts the diversity. The images generated from these noisy initial images exhibited richer background details compared to those generated from the original images, even a small amount of noise can significantly alter the background texture. The diversity of the final images is proportional to the initial image's noise amount, which aligned with our assumption in Section \ref{sec:eq}.

To further explore how adding artificial noise to the initial image affects the diversity of the generated image and background details, Gaussian noise with varying standard deviation \( \sigma \) was introduced at the pixel level, as shown in Figure \ref{fig:noise}.
As a result, Figure \ref{fig:noise} demonstrates the significant impact that noise has on the diversity of the generated images. When noise is introduced into the initial images, even in small amounts, it drastically alters the background texture, resulting in richer and more varied background details compared to images generated from noise-free inputs. This is also validated in the Table \ref{tab:benchmark}. The degree of diversity in the final images is directly proportional to the amount of noise added to the initial image, supporting our hypothesis outlined in Section \ref{sec:eq}.

In addition, we investigated how different types of noise affect the richness and detail of the generated images. We experimented with various types of noise, including Gaussian noise, salt-and-pepper noise, and Perlin noise \cite{green2005implementing}, by applying each to the initial image and measuring the resulting information entropy.
As shown in Figure \ref{fig:type}, even a small amount of artificial signal can enhance background synthesis, and different types of noise lead to varying levels of background detail in the generated images. For instance, coarse noise patterns tend to generate blurred backgrounds, while fine noise results in more intricate textures, as shown in Figure \ref{fig:type}. 

% Meanwhile, over-increasing the noise ratio may causes the denoising network to misinterpret the input, leading to the loss of major image components
% In contrast, salt-and-pepper noise produces images with sharp textures, indicating that these irregular signals introduce a significant amount of noise into the latent space, which facilitates the formation of random patterns. 
% Perlin noise, however, behaves differently. Changes in the noise details do not significantly increase information entropy. 
% Our results show that coarse noise patterns tend to generate blurred backgrounds, while fine noise results in more intricate textures.
% Figure \ref{fig:type} demonstrates that different types of noise lead to varying levels of background detail in the generated images. Gaussian noise typically produces smooth and consistent textures, while salt-and-pepper noise creates more distinct and varied patterns. 

% This observation aligns with our earlier assumption that adding even minimal noise increases the information entropy of the image, which enhances the diffusion process. The noise introduces subtle variations that the diffusion model amplifies during the generation, leading to more diverse outcomes. This validates our previous theoretical insights into the role of noise in improving diversity.

\section{Experiments}

Our experiments focus on evaluating two key aspects of the generated dendrite samples: data diversity and morphology consistency to the input masks. Then, we compare our proposed method (Prism) against other approaches using conventional evaluation metrics.

\subsection{Qualitative Results}
Figure \ref{fig:grid} presents the qualitative results generated by different diffusion frameworks, including standard latent diffusion \cite{rombach2022high}, control-based diffusion methods \cite{zhang2023adding,zhao2024uni}, and the proposed Diffusion Prism. The inputs are binary masks, and the outputs are image-to-image diffusion-generated samples.  
As shown in the figure, vanilla SD1.5 produces realistic samples; however, as the denoising strength increases, significant morphological details are lost. ControlNet, when paired with a specific preprocessor (Canny), achieves better results than other preprocessors, yet the generated backgrounds still lack diversity. Uni-ControlNet improves both diversity and morphological consistency, addressing some of these issues. In comparison, the proposed Diffusion Prism demonstrates superior performance in both visual realism and diversity, achieving the best overall quality among the evaluated methods.

% 3 in 1 Figure
 \begin{figure*}[htbp]
    \centering
    \centerline{\includegraphics[width=1\textwidth]{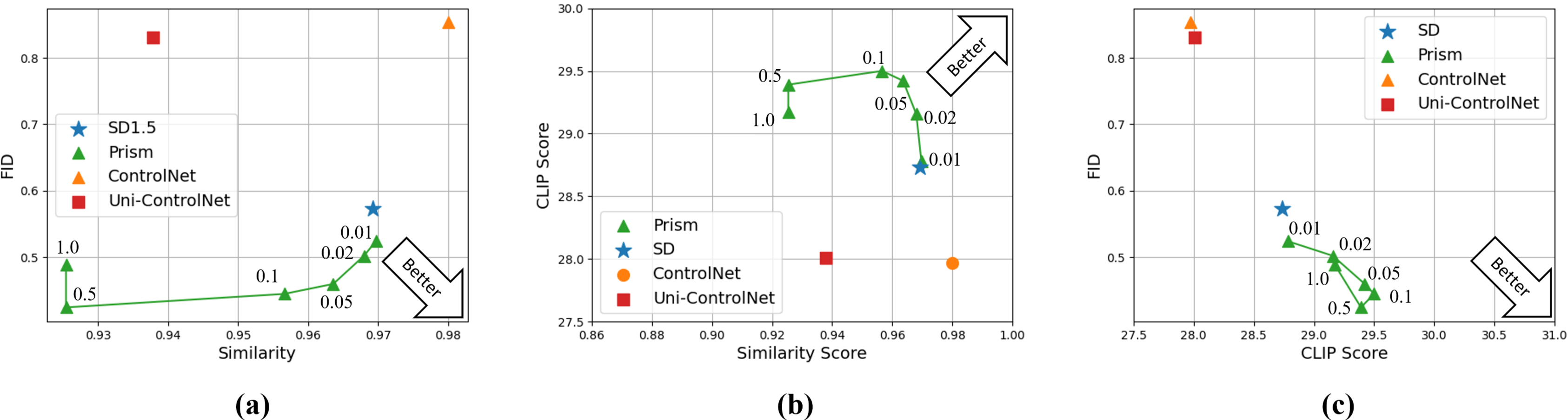}}
    \caption{Model performance comparison. (a) is the nFID-SSIM plot, (b) is the CLIP-SSIM plot, and (c) is the nFID-CLIP plot}
    \label{fig:exp}
    \vspace{-0.25cm}
\end{figure*}

\subsection{Quantitative evaluation}

To quantitatively assess the performance of our method, we conduct comparisons with both vanilla SD1.5 (Stable Diffusion v1.5) \cite{rombach2022high}, ControlNet (SD1.5) \cite{zhang2023adding}, and Uni-ControlNet (SD 1.5)\cite{zhao2024uni}. For fairness, we applied the Prism module to the vanilla SD1.5 and kept the diffusion parameters consistent across all experiments. Specifically,
\begin{itemize}
    \item Denoising steps, refer to the number of iterations in the denoising process, we use $10$ in all experiments.
    \item Denoising strength, controls the balance between noise and the input image during the generation process. We use the value $0.3$ in the vallina SD 1.5 and the proposed Prism experiments to ensure the morphology consistency as explained in Section \ref{sec:eq}, and we use the value $0.99$ in other control-based diffusion methods to ensure maximum variation since their morphology will be less impacted by the denoising strength.
    \item Text prompt: The text prompt used in all cases is set to \textit{"a realistic dendrite sample"} to ensure consistency across different models.
    \item     The text prompt scale controls how strongly the model adheres to the text prompt during image generation, also known as the guidance scale. We use $10$ it in all experiments.

\end{itemize}

% As we mainly measure the ability of our model in diversity, fidelity, and morphology consistency. We applied standard Gaussian noise with $\mu = 0$ in the Prism and set the color sampling to full randomness. We observe the morphology and diversity changes by varying the standard deviation ($\sigma$) from 0 to 1. 

In the proposed Prism method, we apply standard Gaussian noise with a mean of \( \mu = 0 \) and set the color sampling to complete randomness ($I \sim \mathcal{N}$). To analyze the impact on morphology and diversity, we varied the standard deviation \( \sigma \) of the noise from 0 to 1.

\noindent \textbf{Normalized FID (nFID):}  
The \textbf{Fréchet Inception Distance (FID)} is a widely used metric for evaluating the quality and diversity of generated images by comparing their distribution to real images \cite{szegedy2016rethinking,wright2022artfid}. In our experiments, we compute the Normalized FID (nFID) by comparing the images generated using Prism and other diffusion approaches against those from a baseline model (SD 1.5). A lower nFID value indicates that the generated images are more structurally and texturally similar to the ground truth.
We utilize the EMDS-6 dataset \cite{zhao2022emds}, which closely matches the style of the real dendritic patterns, to calculate the nFID \cite{everaert2023diffusion}. The features for this calculation are extracted using the standard Inception model \cite{szegedy2016rethinking} provided in the \texttt{torchvision} package from \texttt{PyTorch}. For the evaluation, we prepared multiple noise settings, generating $10,000$ images for each setting to compute and normalize the FID scores.
 
% The nFID metric is particularly important for evaluating the \textbf{morphological consistency} of the generated images. We aim to generate images that not only exhibit diversity but also retain the key structural features of the original input. By tracking nFID across different experimental settings, we can quantify how well the generated images preserve structural integrity while enhancing texture complexity.

\begin{figure}[htbp]
    \centering
    \centerline{\includegraphics[width=1\columnwidth]{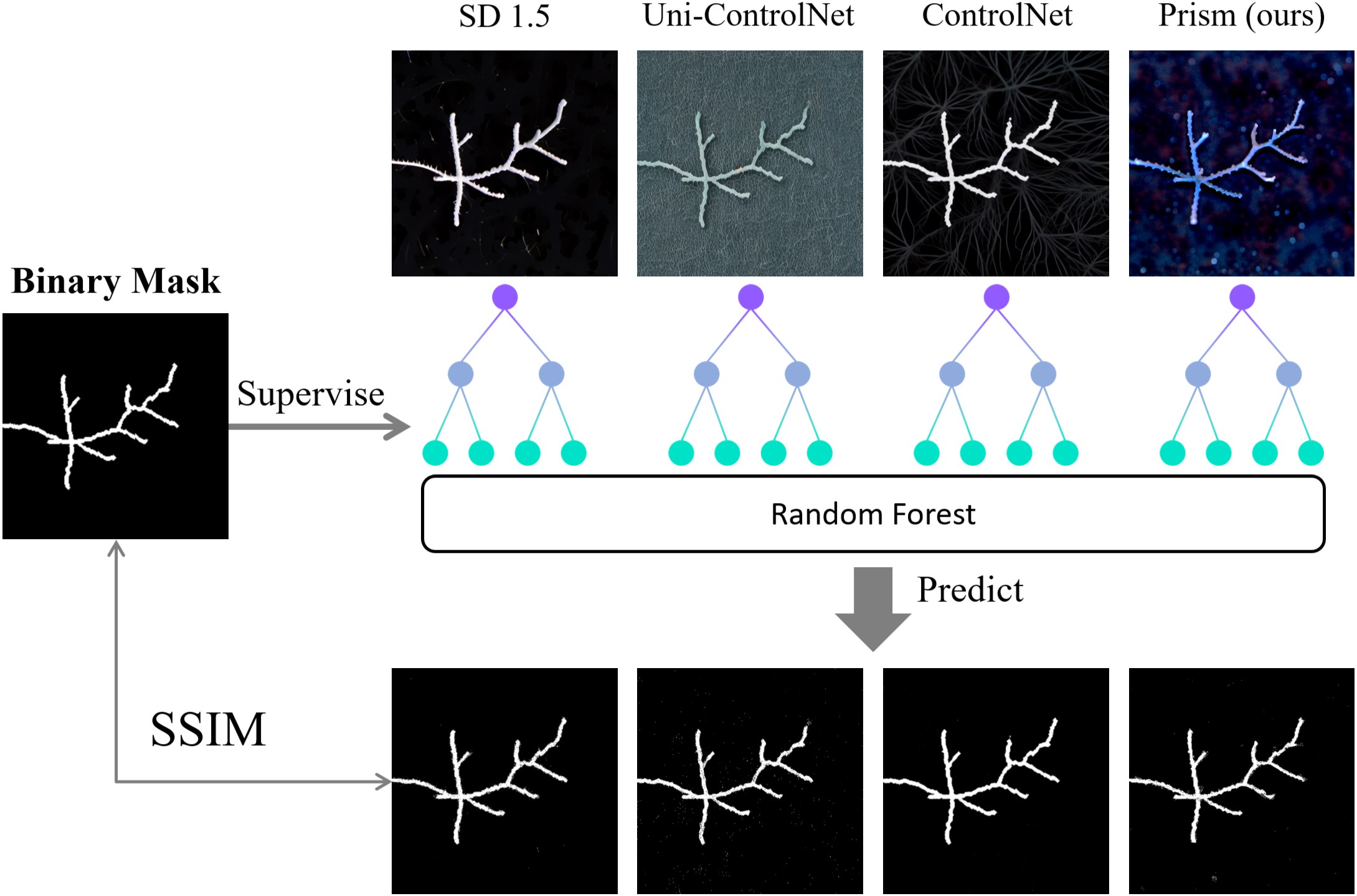}}
    \caption{Process of SSIM calculation. Supervised random forest models will be used to predict the mask of generated samples, then the SSIM will be calculated based on the predictions and ground truth}
    \label{fig:ssim}
    \vspace{-0.25cm}
\end{figure}

 \noindent \textbf{Morphology Similarity: }  To assess morphology consistency, we extract the binary masks from the generated samples using random forest and then compute the structural similarity index measure (SSIM) between the binary mask and the generated image \cite{pinaya2022brain}. This method is reliable and accurate as the original binary masks can be used for ground-truth labels to supervise the random forest model. The SSIM algorithm captures structural similarities and enables the evaluation of how well the generated images preserve the structure of the input. The detailed experiment process is shown in Figure \ref{fig:ssim}.
For convenience, we use the ControlNet with the canny method for comparison due to its better performance, as shown in Figure \ref{fig:grid}.

\noindent \textbf{CLIP Score: }  
Additionally, we measure the text-image similarity through the CLIP Score, which measures how well the generated images align with the textual or visual prompts \cite{radford2021learning}. We use the same CLIP (ViT-B/32) model, which encodes both images and texts into a shared embedding space. A higher CLIP Score indicates that the generated images better align with the semantics of the input prompt.

% In our experiments, we vary the \textbf{strength parameter} to control the level of noise introduced during the denoising process. As the strength increases, more randomness is injected into the diffusion process, potentially leading to more diverse outputs. By using the CLIP Score, we can observe how well this additional diversity translates into semantically relevant images. A key goal is to balance diversity with maintaining the alignment to the original input's semantics.

\begin{table}[htbp]  % Begin table environment
    \centering      % Center the table
    \caption{Performance Comparison
    % \xc{You should split this table into two tables. Keep one of your models with the best performance. The others should go  ablation analysis.}
    }
    \resizebox{1\linewidth}{!}{   \begin{tabular}{lccc} % Specify columns (|c| for centered, |l| left, |r| right)
        \toprule      % Horizontal line 
        
         Method&nFID-10k $\downarrow$ &CLIP Score $\uparrow$&SSIM $\uparrow$\\
        \midrule
         SD1.5 \cite{rombach2022high}&0.6039& 28.73& 0.9692\\
  ControlNet (Best)\cite{zhang2023adding}&0.8531& 27.97&\textbf{0.9801}\\
 Uni-ControlNet\cite{zhao2024uni}& 0.8311& 28.01&0.9378\\
 Fully Random (denoise strength = 1.0)& 0.8893& 27.77&0.5067\\
         \midrule
  Prism (noise Std. = 0.01)&0.5238& 28.78&{0.9697}
\\
  Prism (noise Std. = 0.05)&0.4594& 29.42& 0.9636
\\ 
  Prism (noise Std. = 0.1)&0.4444& \textbf{29.47}& 0.9566
\\
 Prism (noise Std. = 0.5)&  \textbf{0.4241}& 29.40& 0.9254
\\
 Prism (noise Std. = 1.0)& 0.4887& 29.19&0.9254
\\\bottomrule
    \end{tabular}}
    \label{tab:benchmark}             % Add a label for referencing
    \vspace{-0.25cm}
\end{table}

\noindent \textbf{FID-Similarity Test:}  
For reference, we use full denoising strength ($1.0$) to represent the fully random samples, as shown in Table \ref{tab:benchmark}.
As illustrated in Figure \ref{fig:exp}(a), ControlNet achieves the highest CLIP similarity score, indicating that the morphology of its generated images is closely aligned with the input masks. However, its FID score remains high, indicating lower quality. As illustrated in Figure \ref{fig:grid}, although the input image’s structure is well preserved, it does not blend seamlessly with the generated background.

In contrast, the images generated by Prism, with a noise level of $\sigma=0.01$, maintain a comparable SSIM ($0.92-0.97$) while enhancing diversity, as shown in Table \ref{tab:benchmark}. As the noise level increases, the FID continues to improve until $\sigma=0.5$. Beyond this point, excessive noise degrades the input signal in the pixel space, leading to a reduction in both image quality and similarity. 

% Specifically, the FID improves as the noise level increases, reaching its optimal value at \( \sigma = 0.05 \). Beyond this point, further increases in noise result in diminished image quality and higher FID scores.

\noindent \textbf{CLIP Score-Similarity Test:}  
As shown in Figure \ref{fig:exp}(b), Prism-generated images achieve higher CLIP scores due to their diverse backgrounds and realistic dendritic patterns. In comparison, the CLIP score of ControlNet falls below that of vanilla SD 1.5, suggesting that controllable image-to-image diffusion models such as ControlNet and Uni-ControlNet may not be optimal for this particular task.

\noindent \textbf{FID-CLIP Score Test:}  
As demonstrated in Figure \ref{fig:exp}(c), Prism consistently outperforms other methods in both text-image alignment and diversity without significantly compromising image similarity. This balance showcases Prism’s ability to generate visually diverse and semantically consistent images.

\begin{figure}[htbp]
    \centering
    \centerline{\includegraphics[width=1\columnwidth]{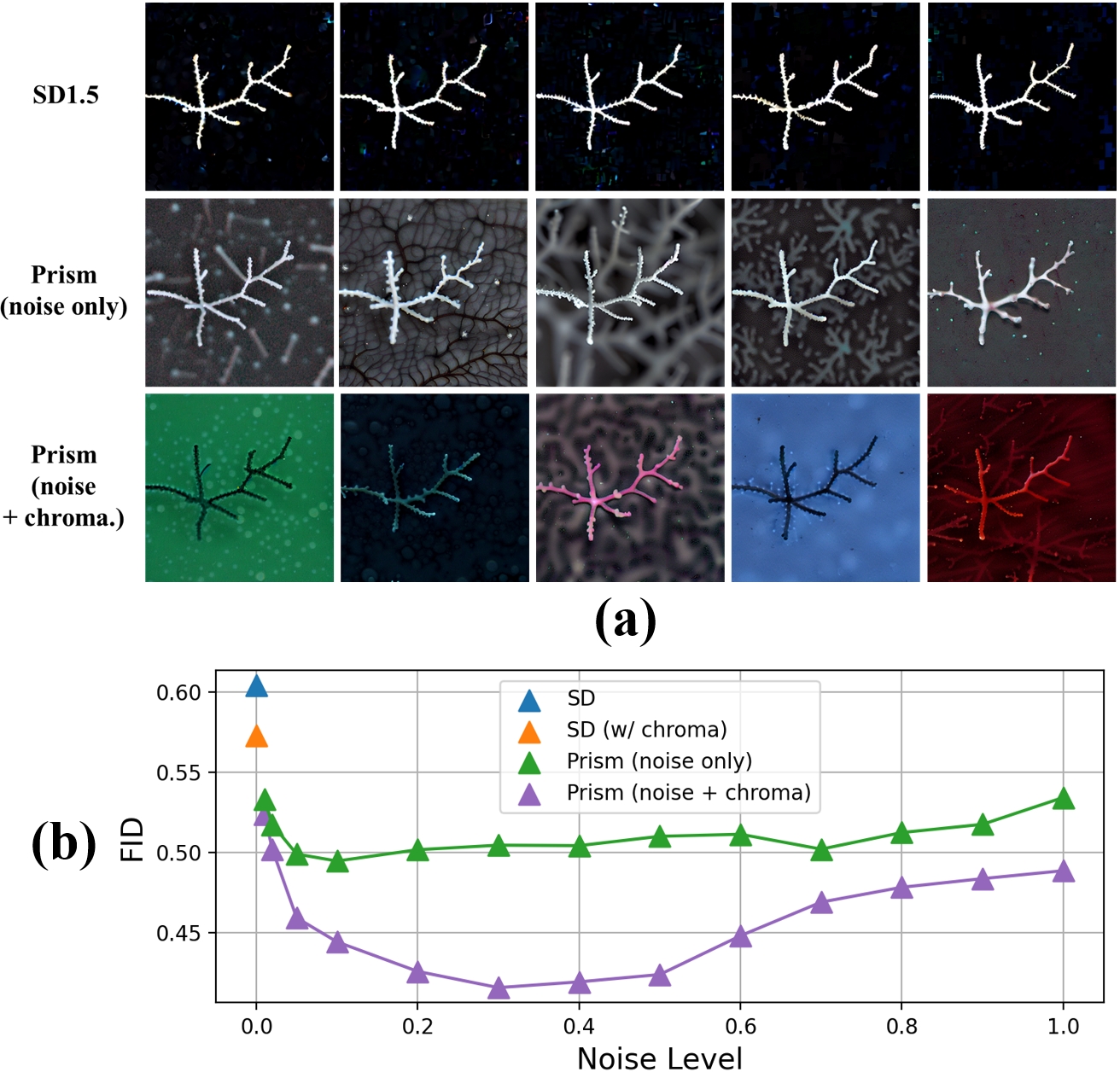}}
    \caption{Ablation study. (a) demonstrate the visual effect of the proposed noise module and chromatic aberration. (b) shows the effect of different modules in terms of nFID score.}
    \label{fig:ablation}
    \vspace{-0.25cm}
\end{figure}

\subsection{Ablation Study}
In this section, we evaluate the two key components of the Prism: controllable noise and chromatic aberration. As shown in Figure \ref{fig:ablation}(b), we conducted tests to measure the FID of generated images under different settings. The results demonstrate that both noise injection and chromatic aberration significantly enhance image quality and diversity.
Notably, the inclusion of chromatic aberration allows for a more effective use of noise, leading to better results compared to the noise-only approach, as shown in Figure \ref{fig:ablation}(a). Table \ref{tab:abl} shows that not only the quality of the image can be improved with both modules, but also they boost the text-image alignment and the morphology consistency. 
The combination of these two techniques results in more diverse and higher-quality outputs, confirming their importance in improving the overall performance of the model.

%  \begin{figure}[htbp]
%     \centering
%     \centerline{\includegraphics[width=0.8\columnwidth]{Figure/ablation_2.png}}
%     \caption{The concept of Mask Diffuser: synthesize realistic samples using ground truth-masks.}
%     \label{fig:ablation}
% \end{figure}

\begin{table}[htbp]  % Begin table environment
    \centering      % Center the table
    \caption{Ablation Study
    % \xc{You should split this table into two tables. Keep one of your models with the best performance. The others should go  ablation analysis.}
    }
    \resizebox{1\linewidth}{!}{   \begin{tabular}{llll} % Specify columns (|c| for centered, |l| left, |r| right)
        \toprule      % Horizontal line 
        
         Method (best)&nFID-10k $\downarrow$ &CLIP Score $\uparrow$&SSIM $\uparrow$\\
        \midrule
         SD1.5&0.6039& 28.73& 0.9692\\
  SD+Prism (noise-only)&0.5330 (\better{-0.0709})& 28.78 (\better{-0.05})&0.9718 (\better{-0.0026})\\
  SD+Prism (chroma-only)&0.5727 (\better{-0.0312})& 29.03 (\better{-0.30})&\textbf{0.9726} (\better{-0.0034})\\
  SD+Prism (noise+chroma)&\textbf{0.4241} (\better{-0.1798})& \textbf{29.47} (\better{-0.74})&0.9697 (\better{-0.0005})\\\bottomrule
    \end{tabular}}
    \label{tab:abl}             % Add a label for referencing
    \vspace{-0.5cm}
\end{table}

\section{Discussion}
\begin{figure}[htbp]
    \centering
    \centerline{\includegraphics[width=1\linewidth]{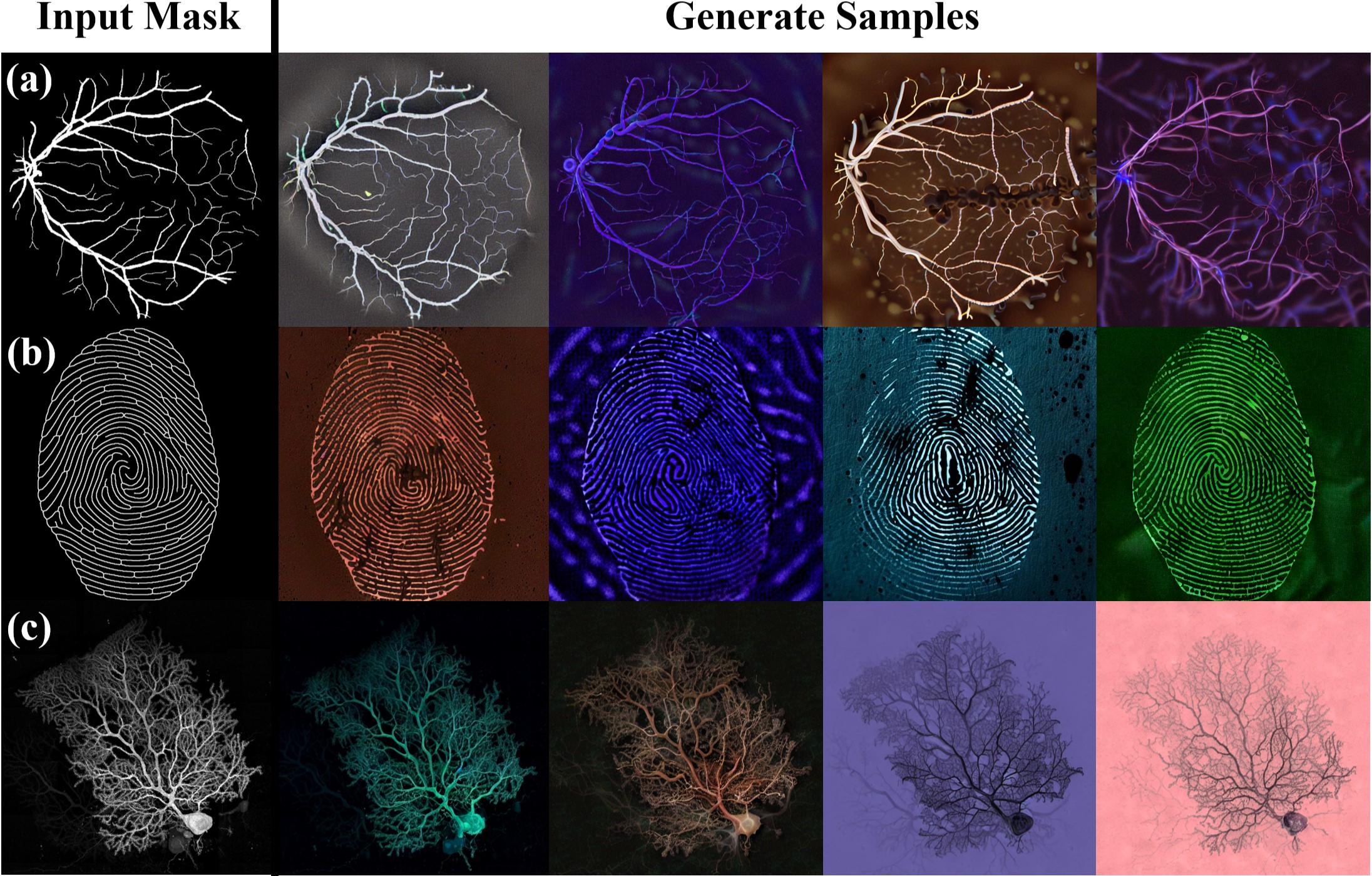}}
    \caption{Diffusion Prism in biometric applications. (a) is Retina Fundus mask with text prompt: \textit{"a realistic retina fundus sample"}, (b) is Fingerprint mask with text prompt: \textit{"a broken fingerprint with ink",} and (c) is Purkinje neuron mask, with text prompt: \textit{"a realistic neuron sample"}}
    \label{fig:bio}
    \vspace{-0.25cm}
\end{figure}

We further evaluated the potential of the proposed Diffusion Prism for data augmentation across other biometric applications. Specifically, Figure \ref{fig:bio} illustrates its application to enhancing binary masks of the retina fundus pattern \cite{1282003}, the fingerprint sample \cite{10216721}, and the Purkinje neuron sample \cite{CCDB2002}. The generated samples present realistic styles and diverse backgrounds while maintaining consistent morphology aligned with the input binary masks, which proved the adaptability and effectiveness in other domains.

\section{Conclusion}

In this paper, we presented the Diffusion Prism, a simple yet effective technique for generating diverse and morphologically consistent images from sparse binary masks. After comprehensively analyzing the signal transmission in image-to-image diffusion, we proposed an effective combination of controlled noise and chromatic aberration to enhance diversity without sacrificing the structural integrity of the input masks. The experimental results on dendritic patterns demonstrated that our method significantly improves the diversity of generated images, outperforming baseline and other methods in both quantitative evaluations and visual comparisons. We also performed the method on other biometric samples, which offers a promising mask-to-image solution in various practical applications.

\section*{Acknowledgments}
This material is based upon the work supported by the National Science Foundation under Grant Number 2204721 and partially supported by our collaborative project with MIT Lincoln Lab under Grant Number 7000612889. 

The authors would like to thank Dr. Siyu Huang for his comments on experiment designs. 

% This work is supported in part by the USDA AFRI program under grant \#2020-67017-33078
% and the National Science Foundation EPSCoR Program under NSF Award \#OIA-2242812.

%%%%%%%%% REFERENCES
{\small
\bibliographystyle{ieee_fullname}
\bibliography{egbib}

\begin{thebibliography}{10}\itemsep=-1pt

\bibitem{bonaldi2023deep}
Lorenza Bonaldi, Andrea Pretto, Carmelo Pirri, Francesca Uccheddu, Chiara~Giulia Fontanella, and Carla Stecco.
\newblock Deep learning-based medical images segmentation of musculoskeletal anatomical structures: a survey of bottlenecks and strategies.
\newblock {\em Bioengineering}, 10(2):137, 2023.

\bibitem{chen2023dh}
Xiwen Chen, Hao Wang, Abolfazl Razi, Michael Kozicki, and Christopher Mann.
\newblock Dh-gan: a physics-driven untrained generative adversarial network for holographic imaging.
\newblock {\em Optics Express}, 31(6):10114--10135, 2023.

\bibitem{chen2024enhancing}
Xiwen Chen, Hao Wang, Zhao Zhang, Zhenmin Li, Huayu Li, Tong Ye, and Abolfazl Razi.
\newblock Enhancing digital hologram reconstruction using reverse-attention loss for untrained physics-driven deep learning models with uncertain distance.
\newblock In {\em AI and Optical Data Sciences V}, volume 12903, pages 132--141. SPIE, 2024.

\bibitem{choi2023custom}
Jooyoung Choi, Yunjey Choi, Yunji Kim, Junho Kim, and Sungroh Yoon.
\newblock Custom-edit: Text-guided image editing with customized diffusion models.
\newblock {\em arXiv preprint arXiv:2305.15779}, 2023.

\bibitem{cuntz2007optimization}
Hermann Cuntz, Alexander Borst, and Idan Segev.
\newblock Optimization principles of dendritic structure.
\newblock {\em Theoretical Biology and Medical Modelling}, 4:1--8, 2007.

\bibitem{everaert2023diffusion}
Martin~Nicolas Everaert, Marco Bocchio, Sami Arpa, Sabine S{\"u}sstrunk, and Radhakrishna Achanta.
\newblock Diffusion in style.
\newblock In {\em Proceedings of the IEEE/CVF International Conference on Computer Vision}, pages 2251--2261, 2023.

\bibitem{everaert2024exploiting}
Martin~Nicolas Everaert, Athanasios Fitsios, Marco Bocchio, Sami Arpa, Sabine S{\"u}sstrunk, and Radhakrishna Achanta.
\newblock Exploiting the signal-leak bias in diffusion models.
\newblock In {\em Proceedings of the IEEE/CVF Winter Conference on Applications of Computer Vision}, pages 4025--4034, 2024.

\bibitem{fang2024fairness}
Meiling Fang, Wufei Yang, Arjan Kuijper, Vitomir Struc, and Naser Damer.
\newblock Fairness in face presentation attack detection.
\newblock {\em Pattern Recognition}, 147:110002, 2024.

\bibitem{green2005implementing}
Simon Green.
\newblock Implementing improved perlin noise.
\newblock {\em GPU Gems}, 2:409--416, 2005.

\bibitem{holden2017fast}
Daniel Holden, Ikhsanul Habibie, Ikuo Kusajima, and Taku Komura.
\newblock Fast neural style transfer for motion data.
\newblock {\em IEEE computer graphics and applications}, 37(4):42--49, 2017.

\bibitem{kawano2024maskdiffusion}
Yasufumi Kawano and Yoshimitsu Aoki.
\newblock Maskdiffusion: Exploiting pre-trained diffusion models for semantic segmentation.
\newblock {\em arXiv preprint arXiv:2403.11194}, 2024.

\bibitem{kirillov2023segment}
Alexander Kirillov, Eric Mintun, Nikhila Ravi, Hanzi Mao, Chloe Rolland, Laura Gustafson, Tete Xiao, Spencer Whitehead, Alexander~C Berg, Wan-Yen Lo, et~al.
\newblock Segment anything.
\newblock {\em arXiv preprint arXiv:2304.02643}, 2023.

\bibitem{lahiri2020retinal}
Avisek Lahiri, Vineet Jain, Arnab Mondal, and Prabir~Kumar Biswas.
\newblock Retinal vessel segmentation under extreme low annotation: A gan based semi-supervised approach.
\newblock In {\em 2020 IEEE international conference on image processing (ICIP)}, pages 418--422. IEEE, 2020.

\bibitem{lee2024microstructure}
Kang-Hyun Lee and Gun~Jin Yun.
\newblock Microstructure reconstruction using diffusion-based generative models.
\newblock {\em Mechanics of Advanced Materials and Structures}, 31(18):4443--4461, 2024.

\bibitem{lin2021temimagenet}
Ruoqian Lin, Rui Zhang, Chunyang Wang, Xiao-Qing Yang, and Huolin~L Xin.
\newblock Temimagenet training library and atomsegnet deep-learning models for high-precision atom segmentation, localization, denoising, and deblurring of atomic-resolution images.
\newblock {\em Scientific reports}, 11(1):5386, 2021.

\bibitem{loey2020deep}
Mohamed Loey, Gunasekaran Manogaran, and Nour Eldeen~M Khalifa.
\newblock A deep transfer learning model with classical data augmentation and cgan to detect covid-19 from chest ct radiography digital images.
\newblock {\em Neural Computing and Applications}, pages 1--13, 2020.

\bibitem{ma2024segment}
Jun Ma, Yuting He, Feifei Li, Lin Han, Chenyu You, and Bo Wang.
\newblock Segment anything in medical images.
\newblock {\em Nature Communications}, 15(1):654, 2024.

\bibitem{CCDB2002}
Martone Maryann, Price Diana, Thor Andrea, Terada Masako, and Hakozaki Hiro.
\newblock mus musculus, purkinje neuron. cil. dataset., 2002.
\newblock CCDB:3687.

\bibitem{melnik2024face}
Andrew Melnik, Maksim Miasayedzenkau, Dzianis Makaravets, Dzianis Pirshtuk, Eren Akbulut, Dennis Holzmann, Tarek Renusch, Gustav Reichert, and Helge Ritter.
\newblock Face generation and editing with stylegan: A survey.
\newblock {\em IEEE Transactions on Pattern Analysis and Machine Intelligence}, 2024.

\bibitem{meng2021sdedit}
Chenlin Meng, Yutong He, Yang Song, Jiaming Song, Jiajun Wu, Jun-Yan Zhu, and Stefano Ermon.
\newblock Sdedit: Guided image synthesis and editing with stochastic differential equations.
\newblock {\em arXiv preprint arXiv:2108.01073}, 2021.

\bibitem{minaee2023biometrics}
Shervin Minaee, Amirali Abdolrashidi, Hang Su, Mohammed Bennamoun, and David Zhang.
\newblock Biometrics recognition using deep learning: A survey.
\newblock {\em Artificial Intelligence Review}, 56(8):8647--8695, 2023.

\bibitem{mokady2023null}
Ron Mokady, Amir Hertz, Kfir Aberman, Yael Pritch, and Daniel Cohen-Or.
\newblock Null-text inversion for editing real images using guided diffusion models.
\newblock In {\em Proceedings of the IEEE/CVF Conference on Computer Vision and Pattern Recognition}, pages 6038--6047, 2023.

\bibitem{ozbey2023unsupervised}
Muzaffer {\"O}zbey, Onat Dalmaz, Salman~UH Dar, Hasan~A Bedel, {\c{S}}aban {\"O}zturk, Alper G{\"u}ng{\"o}r, and Tolga {\c{C}}ukur.
\newblock Unsupervised medical image translation with adversarial diffusion models.
\newblock {\em IEEE Transactions on Medical Imaging}, 2023.

\bibitem{pinaya2022brain}
Walter~HL Pinaya, Petru-Daniel Tudosiu, Jessica Dafflon, Pedro~F Da~Costa, Virginia Fernandez, Parashkev Nachev, Sebastien Ourselin, and M~Jorge Cardoso.
\newblock Brain imaging generation with latent diffusion models.
\newblock In {\em MICCAI Workshop on Deep Generative Models}, pages 117--126. Springer, 2022.

\bibitem{radford2021learning}
Alec Radford, Jong~Wook Kim, Chris Hallacy, Aditya Ramesh, Gabriel Goh, Sandhini Agarwal, Girish Sastry, Amanda Askell, Pamela Mishkin, Jack Clark, et~al.
\newblock Learning transferable visual models from natural language supervision.
\newblock In {\em International conference on machine learning}, pages 8748--8763. PMLR, 2021.

\bibitem{rombach2022high}
Robin Rombach, Andreas Blattmann, Dominik Lorenz, Patrick Esser, and Bj{\"o}rn Ommer.
\newblock High-resolution image synthesis with latent diffusion models.
\newblock In {\em Proceedings of the IEEE/CVF conference on computer vision and pattern recognition}, pages 10684--10695, 2022.

\bibitem{song2020denoising}
Jiaming Song, Chenlin Meng, and Stefano Ermon.
\newblock Denoising diffusion implicit models.
\newblock {\em arXiv preprint arXiv:2010.02502}, 2020.

\bibitem{spruston2008pyramidal}
Nelson Spruston.
\newblock Pyramidal neurons: dendritic structure and synaptic integration.
\newblock {\em Nature Reviews Neuroscience}, 9(3):206--221, 2008.

\bibitem{1282003}
J. Staal, M.D. Abramoff, M. Niemeijer, M.A. Viergever, and B. van Ginneken.
\newblock Ridge-based vessel segmentation in color images of the retina.
\newblock {\em IEEE Transactions on Medical Imaging}, 23(4):501--509, 2004.

\bibitem{szegedy2016rethinking}
Christian Szegedy, Vincent Vanhoucke, Sergey Ioffe, Jon Shlens, and Zbigniew Wojna.
\newblock Rethinking the inception architecture for computer vision.
\newblock In {\em Proceedings of the IEEE conference on computer vision and pattern recognition}, pages 2818--2826, 2016.

\bibitem{torkzadehmahani2019dp}
Reihaneh Torkzadehmahani, Peter Kairouz, and Benedict Paten.
\newblock Dp-cgan: Differentially private synthetic data and label generation.
\newblock In {\em Proceedings of the IEEE/CVF Conference on Computer Vision and Pattern Recognition Workshops}, pages 0--0, 2019.

\bibitem{wang2024flame}
Hao Wang, Sayed Pedram~Haeri Boroujeni, Xiwen Chen, Ashish Bastola, Huayu Li, Wenhui Zhu, and Abolfazl Razi.
\newblock Flame diffuser: Wildfire image synthesis using mask guided diffusion.
\newblock {\em arXiv preprint arXiv:2403.03463}, 2024.

\bibitem{10216721}
Hao Wang, Xiwen Chen, Abolfazl Razi, and Rahul Amin.
\newblock { Fast Key Points Detection and Matching for Tree-Structured Images }.
\newblock In {\em 2022 International Conference on Computational Science and Computational Intelligence (CSCI)}, pages 1381--1387, Los Alamitos, CA, USA, Dec. 2022. IEEE Computer Society.

\bibitem{10216773}
Hao Wang, Xiwen Chen, Abolfazl Razi, Michael Kozicki, Rahul Amin, and Mark Manfredo.
\newblock Nano-resolution visual identifiers enable secure monitoring in next-generation cyber-physical systems.
\newblock In {\em 2022 International Conference on Computational Science and Computational Intelligence (CSCI)}, pages 856--861, 2022.

\bibitem{wang2024rbad}
Hao Wang, Wenhui Zhu, Jiayou Qin, Xin Li, Oana Dumitrascu, Xiwen Chen, Peijie Qiu, and Abolfazl Razi.
\newblock Rbad: A dataset and benchmark for retinal vessels branching angle detection.
\newblock {\em arXiv preprint arXiv:2407.12271}, 2024.

\bibitem{wright2022artfid}
Matthias Wright and Bj{\"o}rn Ommer.
\newblock Artfid: Quantitative evaluation of neural style transfer.
\newblock In {\em DAGM German Conference on Pattern Recognition}, pages 560--576. Springer, 2022.

\bibitem{wu2019u}
Cong Wu, Yixuan Zou, and Zhi Yang.
\newblock U-gan: Generative adversarial networks with u-net for retinal vessel segmentation.
\newblock In {\em 2019 14th international conference on computer science \& education (ICCSE)}, pages 642--646. IEEE, 2019.

\bibitem{wu2024medsegdiff}
Junde Wu, Wei Ji, Huazhu Fu, Min Xu, Yueming Jin, and Yanwu Xu.
\newblock Medsegdiff-v2: Diffusion-based medical image segmentation with transformer.
\newblock In {\em Proceedings of the AAAI Conference on Artificial Intelligence}, volume~38, pages 6030--6038, 2024.

\bibitem{zhang2024transparent}
Lvmin Zhang and Maneesh Agrawala.
\newblock Transparent image layer diffusion using latent transparency.
\newblock {\em arXiv preprint arXiv:2402.17113}, 2024.

\bibitem{zhang2023adding}
Lvmin Zhang, Anyi Rao, and Maneesh Agrawala.
\newblock Adding conditional control to text-to-image diffusion models.
\newblock In {\em Proceedings of the IEEE/CVF International Conference on Computer Vision}, pages 3836--3847, 2023.

\bibitem{zhao2022emds}
Peng Zhao, Chen Li, Md~Mamunur Rahaman, Hao Xu, Pingli Ma, Hechen Yang, Hongzan Sun, Tao Jiang, Ning Xu, and Marcin Grzegorzek.
\newblock Emds-6: Environmental microorganism image dataset sixth version for image denoising, segmentation, feature extraction, classification, and detection method evaluation.
\newblock {\em Frontiers in Microbiology}, 13:829027, 2022.

\bibitem{zhao2024uni}
Shihao Zhao, Dongdong Chen, Yen-Chun Chen, Jianmin Bao, Shaozhe Hao, Lu Yuan, and Kwan-Yee~K Wong.
\newblock Uni-controlnet: All-in-one control to text-to-image diffusion models.
\newblock {\em Advances in Neural Information Processing Systems}, 36, 2024.

\bibitem{zhou2023denoising}
Linqi Zhou, Aaron Lou, Samar Khanna, and Stefano Ermon.
\newblock Denoising diffusion bridge models.
\newblock {\em arXiv preprint arXiv:2309.16948}, 2023.

\bibitem{zhou2023maskdiffusion}
Yupeng Zhou, Daquan Zhou, Zuo-Liang Zhu, Yaxing Wang, Qibin Hou, and Jiashi Feng.
\newblock Maskdiffusion: Boosting text-to-image consistency with conditional mask.
\newblock {\em arXiv preprint arXiv:2309.04399}, 2023.

\bibitem{zhu2023beyond}
Wenhui Zhu, Peijie Qiu, Xiwen Chen, Huayu Li, Hao Wang, Natasha Lepore, Oana~M Dumitrascu, and Yalin Wang.
\newblock Beyond mobilenet: An improved mobilenet for retinal diseases.
\newblock In {\em International Conference on Medical Image Computing and Computer-Assisted Intervention}, pages 56--65. Springer, 2023.

\bibitem{zhu2023otre}
Wenhui Zhu, Peijie Qiu, Oana~M Dumitrascu, Jacob~M Sobczak, Mohammad Farazi, Zhangsihao Yang, Keshav Nandakumar, and Yalin Wang.
\newblock Otre: where optimal transport guided unpaired image-to-image translation meets regularization by enhancing.
\newblock In {\em International Conference on Information Processing in Medical Imaging}, pages 415--427. Springer, 2023.

\end{thebibliography}
}

\end{document}